%% file: main.tex
\documentclass[10pt,twocolumn,letterpaper]{article}

\usepackage{cvpr}              

\usepackage{graphicx}
\usepackage{amsmath}
\usepackage{amssymb}
\usepackage{booktabs}
\usepackage{arydshln}
\usepackage{bm}
\usepackage{multirow}

\newcommand*{\affaddr}[1]{#1} 
\newcommand*{\affmark}[1][*]{\textsuperscript{#1}}

\renewcommand{\thefootnote}{\fnsymbol{footnote}}

\usepackage[pagebackref,breaklinks,colorlinks]{hyperref}

\usepackage[capitalize]{cleveref}
\crefname{section}{Sec.}{Secs.}
\Crefname{section}{Section}{Sections}
\Crefname{table}{Table}{Tables}
\crefname{table}{Tab.}{Tabs.}

\def\cvprPaperID{10530} 
\def\confName{CVPR}
\def\confYear{2023}

\begin{document}

\title{3D GAN Inversion with Facial Symmetry Prior}


\author{%
Fei Yin\affmark[1], 
Yong Zhang\affmark[2]\footnotemark[2], 
Xuan Wang\affmark[3], 
Tengfei Wang\affmark[4], 
Xiaoyu Li\affmark[2],
Yuan Gong\affmark[1],\\
Yanbo Fan\affmark[2],
Xiaodong Cun\affmark[2],
Ying Shan\affmark[2],
Cengiz Öztireli\affmark[5],
Yujiu Yang\affmark[1]\footnotemark[2],\\
\affaddr{\affmark[1] Shenzhen International Graduate School, Tsinghua University} \\
\affaddr{\affmark[2]Tencent AI Lab \textcolor{white}{spa}}
\affaddr{\affmark[3]Ant Group \textcolor{white}{spa}}
\affaddr{\affmark[4] HKUST \textcolor{white}{spa}}
\affaddr{\affmark[5]University of Cambridge}\\
}
\maketitle
\renewcommand{\thefootnote}{\fnsymbol{footnote}}
\footnotetext{Work done during an internship at Tencent AI Lab.}
\footnotetext[2]{Corresponding Author.}

\begin{abstract}
   \input{tex/abstract.tex}
\end{abstract}


\input{tex/introduction.tex}

\input{tex/relatedwork.tex}

\input{tex/method.tex}

\input{tex/experiment.tex}

\input{tex/conclusion.tex}

{\small
\bibliographystyle{ieee_fullname}
\bibliography{egbib}
}

\input{tex/appendix}
\end{document}

%% file: tex/abstract.tex
%
Recently, a surge of high-quality 3D-aware GANs have been proposed, which leverage the generative power of neural rendering.
It is natural to associate 3D GANs with GAN inversion methods to project a real image into the generator's latent space, allowing free-view consistent synthesis and editing, referred as 3D GAN inversion.
Although with the facial prior preserved in pre-trained 3D GANs, reconstructing a 3D portrait with only one monocular image is still an ill-pose problem.
The straightforward application of 2D GAN inversion methods focuses on texture similarity only while ignoring the correctness of 3D geometry shapes.
It may raise geometry collapse effects, especially when reconstructing a side face under an extreme pose.
Besides, the synthetic results in novel views are prone to be blurry.
In this work, we propose a novel method to promote 3D GAN inversion by introducing facial symmetry prior.
We design a pipeline and constraints to make full use of the pseudo auxiliary view obtained via image flipping, which helps obtain a view-consistent and well-structured geometry shape during the inversion process.
To enhance texture fidelity in unobserved viewpoints, pseudo labels from depth-guided 3D warping can provide extra supervision.
We design constraints to filter out conflict areas for optimization in asymmetric situations.
Comprehensive quantitative and qualitative evaluations on image reconstruction and editing demonstrate the superiority of our method.

%% file: tex/introduction.tex
\section{Introduction}
\label{sec:introduction}

\input{image_tex/img_teaser}



Recent 3D-aware generative adversarial networks~(3D GANs) have seen immense progress. 
By incorporating a neural rendering engine into the generator network architecture, 3D GANs can synthesize view-consistent images.
To increase the generation resolution, existing methods~\cite{chan2022eg3d,or2022stylesdf,gu2021stylenerf,xu20223volumegan,xue2022giraffehd,zhang2022multi,tan2022voluxgan,sun2022ide3d,xiang2022gramhd} boost the 3D inductive bias with an additional 2D CNN-based upsampler or an efficient 3D representation modeling method. 
With tremendous effort, 3D GANs can produce photorealistic images while enforcing strong 3D consistency across different views.

We are interested in the task of reconstructing a human face with 3D geometry and texture given only one monocular image. 
It is an ill-posed problem and close to the harsh condition of real scenarios.
With the power of 3D GANs, it seems achievable via projecting a target image onto the manifold of a pre-trained generator. 
The process is referred as 3D GAN inversion.
A straightforward path is to follow the 2D GAN inversion method~\cite{roich2021PTI}, \textit{i.e.,} optimizing the latent code and the network parameters of the generator to overfit the specific portrait.

However, since the ground truth 3D geometry is absent given one monocular image, the inversion result is far from satisfactory.
The process of fitting a 3D GAN to one image would sacrifice geometric correctness in order to make the synthetic texture as close as possible to the input, even destroying the original semantic-rich latent space.
As the optimization process goes, the face geometry tends to degenerate into a flattened shape, due to the absence of geometry supervision, \textit{e.g.,} images from other views.
Besides, there exist quality issues in texture synthesis under novel views. 
The rendered images of unseen views tend to be blurry and inconsistent with the original image, especially when reconstructing a side face under an extreme pose.  
Because there is no texture supervision for unseen views given only one monocular image. 
The failure cases of directly applying~\cite{roich2021PTI} are illustrated in Fig.~\ref{fig:teaser}.

In this work, to alleviate the issue caused by missing geometry and texture supervision under multiple views, we propose a novel 3D GAN inversion approach by taking full advantage of facial symmetry prior to construct pseudo supervision of different views.   
Intuitively, we note that human faces are almost symmetric. 
Assuming the given portrait is symmetric, we can obtain an additional perspective of the portrait by simply mirroring the image. 
The images of two distinct views can provide geometric relations between the 3D points and their 2D projections based on epipolar geometry.
Motivated by this, we seek to leverage facial symmetry as the geometric prior constraining the inversion. 
The symmetry prior is also employed in a traditional 3D reconstruction work~\cite{wu2020unsupervised}.
We leverage the mirrored image as extra supervision of another view when performing the inversion, which prevents the geometry collapse.
A rough geometry can be obtained by the inversion with the original and mirror images. 

To further enhance texture quality and geometry in novel views, we employ depth-guided 3D warping to generate the pseudo images of the views surrounding the input and symmetric camera pose. 
The depth is inferred from the rough 3D volume. 
The original image along with the pseudo images are used to fine-tune the generator's parameters for the joint promotion of texture and geometry. 
To prevent the optimized geometry from deviating too much from the rough geometry, we design a geometry regularization term as a constraint. 
However, human faces are never fully symmetric in practice, neither in shape nor appearance. 
Therefore, we design several constraints to extract meaningful information adaptively from the mirror image without compromising the original reconstruction quality.

Our main contributions are as follows:

\begin{itemize}
    \item We propose a novel 3D GAN inversion method by incorporating facial symmetry prior. 
    It enables a high-quality reconstruction while preserving the multi-view consistency in geometry and texture. 
    
    \item We conduct comprehensive experiments to demonstrate the effectiveness of our method and compare it with many \textit{state-of-the-art} inversion methods.
    We also apply our method to various downstream applications.
\end{itemize}

%% file: image_tex/img_teaser.tex
\begin{figure}[t]
\begin{center}
\centerline{\includegraphics[width=1\linewidth]{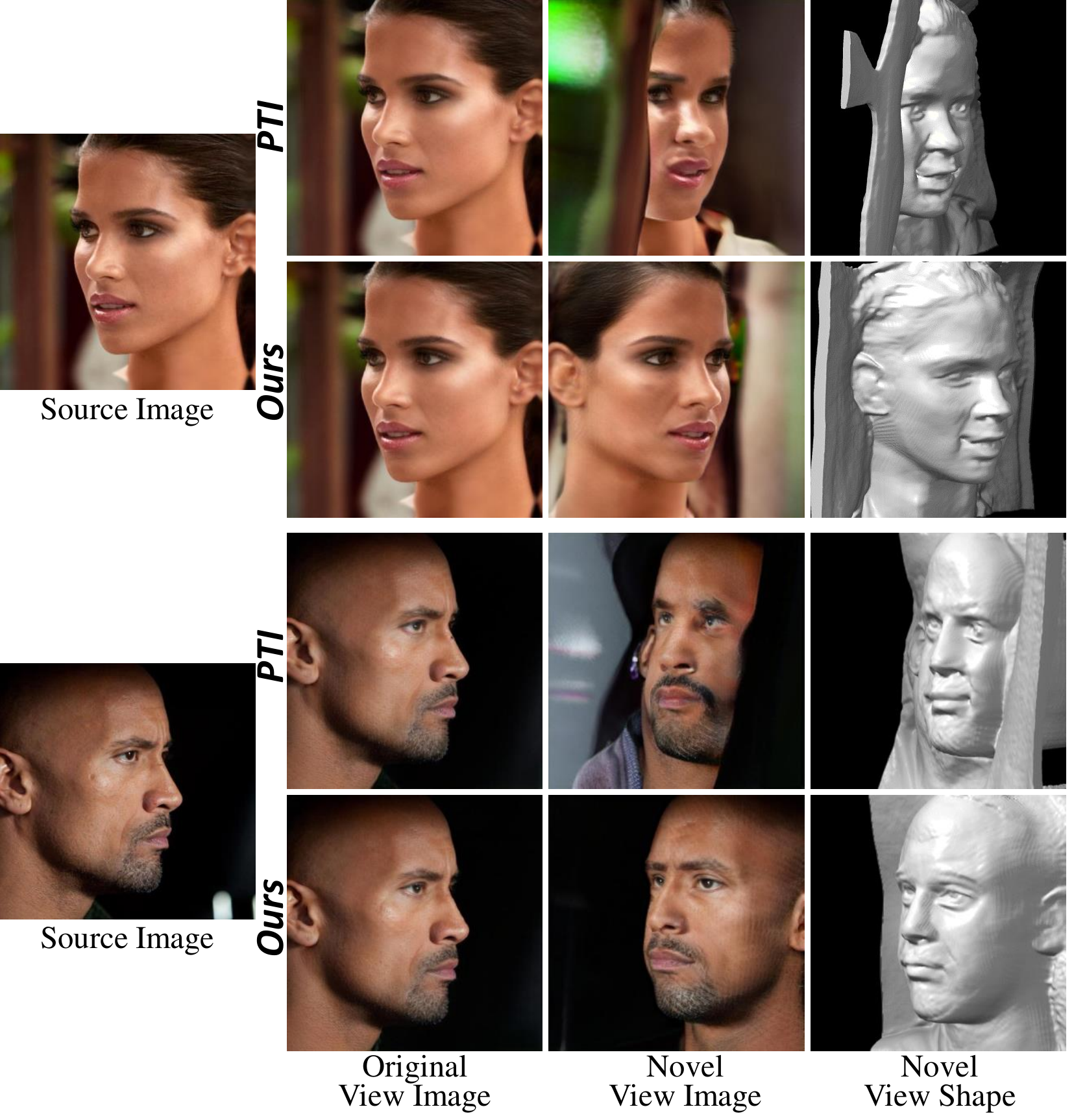}}
\caption{ Visual examples of our inversion method. 
Direct applying 2D GAN inversion methods~(PTI~\cite{roich2021PTI}) to the 3D GAN suffers from inaccurate geometry in novel views.
Our method excels in synthesizing consistent geometry  and high-fidelity texture in different views, even reconstructing a face under an extreme pose.
}
\label{fig:teaser}
\vspace{-1.2cm}
\end{center}
\end{figure}

%% file: tex/relatedwork.tex
\section{Related Work}
\label{sec:related_work}

\subsection{3D-Aware GANs}
Recently, neural scene representations have incorporated 3D prior into image synthesis with explicit camera control.
Inspired by the success of Neural Radiance Fields~(NeRF)~\cite{mildenhall2021nerf}, \cite{chan2021pigan,niemeyer2021giraffe} employ implicit volumetric neural rendering structure for consistent novel view synthesis, required only unconstrained monocular images training.
To overcome the computational cost and lift the generation resolution, the following methods adopt a two-stage rendering process~\cite{chan2022eg3d,or2022stylesdf,gu2021stylenerf,xu20223volumegan,mejjati2021gaussigan,xue2022giraffehd,zhang2022multi,tan2022voluxgan,sun2022ide3d,zhou2021cips}.
Since 2D upsamplers may introduce view-inconsistent artifacts, NeRF path regularization~\cite{gu2021stylenerf} and dual discriminators~\cite{chan2022eg3d} are proposed.
Different 3D modeling representations are further designed for scalable and fast rendering. EG3D~\cite{chan2022eg3d} introduces tri-plane representation, and GRAM-HD~\cite{xiang2022gramhd} proposes to render radiance manifolds first for efficient sampling.
Boosting with the powerful high-fidelity unconditioned 3D GANs, we can achieve real image 3D reconstruction and editing.
Specifically, we select the \textit{state-of-the-art} EG3D~\cite{chan2022eg3d} as our backbone.

\subsection{GAN Inversion}
To edit a real image~\cite{shen2020interfacegan,yin2022styleheat}, GAN inversion is applied first to discover a corresponding latent code from which the generator can synthesize the real image.
Existing 2D GAN inversion approaches can be categorized into optimization-based, learning-based, and hybrid methods.
\cite{stylegan2,abdal2019image2stylegan} directly minimize the reconstruction distance via optimizing the latent codes.
Learning-based methods~\cite{tov2021e4e,wang2021hfgi,alaluf2021hyperstyle,bai2022padinv} exploit a general encoder network to map the input image into latent space in real-time.
Hybrid methods would apply the latent code predicted from the encoder as initialization in the later optimization process.
Beyond the original inversion latent space, PTI~\cite{roich2021PTI} further optimizes the parameters of the generator to enhance the visual fidelity.

As for the 3D GAN inversion task, most methods directly transfer the 2D methods, \textit{e.g.}, PTI~\cite{roich2021PTI} and e4e~\cite{tov2021e4e}, which may suffer from the poor results in novel views.
Pix2NeRF~\cite{cai2022pix2nerf} introduced a joint distillation strategy for training a 3D inversion encoder.
A concurrent work~\cite{ko20223dganinversionwposeoptimization} proposes to perform camera pose optimization simultaneously to ensure view consistency.
However, none of the above methods take geometry shape into consideration.

\input{image_tex/img_pipeline}

\subsection{Few-shot NeRF}
Few-shot NeRF aims at reconstructing general 3D scenarios where only a few observed views are available, which shares a similar setting with 3D GAN inversion.
MVSNeRF~\cite{chen2021mvsnerf} leverages plane-swept cost volumes in multi-view stereo for geometry-aware scene reasoning to improve performance.
DietNeRF~\cite{jain2021dietnerf} enforces semantic consistency between rendered images from unseen view and seen images via a CLIP encoder~\cite{radford2021clip}. 
RegNeRF~\cite{niemeyer2022regnerf} regularizes the texture of patches rendered from unobserved viewpoints without relying on additional training modules. Since it is hard to find a common prior for general scenes, these methods investigate how to ensure the geometry consistency of different views, which gives us inspiration.

%% file: image_tex/img_pipeline.tex
\begin{figure*}[t]
\begin{center}
\centerline{\includegraphics[width=1\linewidth]{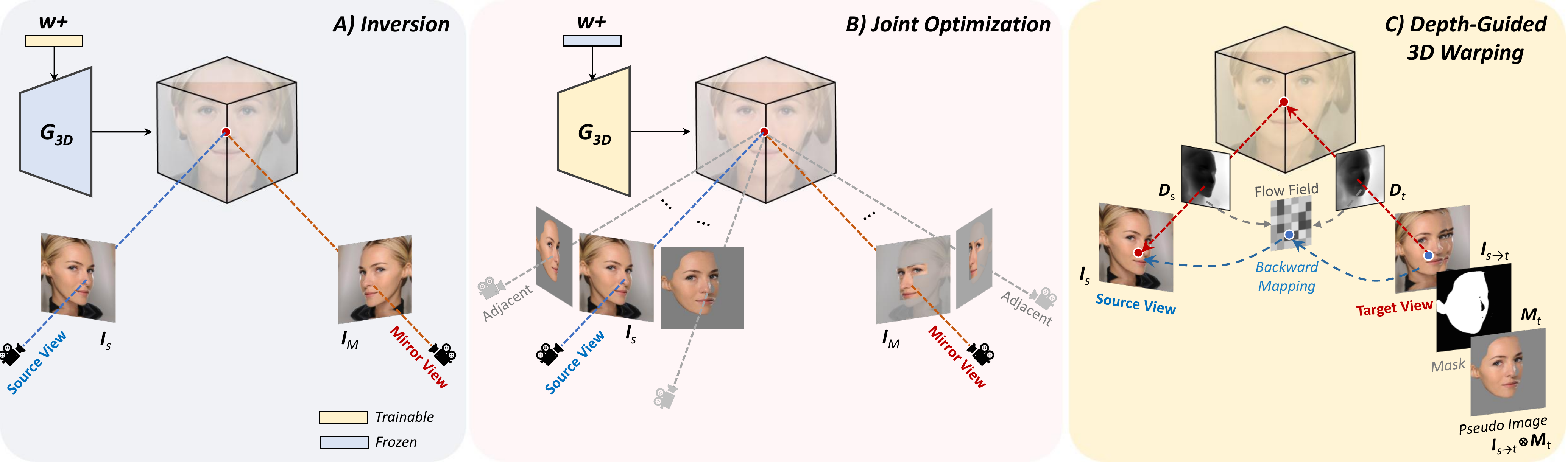}}
\caption{
The proposed framework. \textit{A}) Our method first performs inversion with the help of the symmetry view to achieve the latent code $w^+$ with a roughly correct geometry.
\textit{B}) The original image and the mirror one, along with adjacent warping pseudos, are used for joint optimization to enhance the geometry and texture of rendered images in novel views.
\textit{C}) Depth-guided 3D warping are used to generate pseudo images in novel views to provide extra supervision. Unfaithful regions are filtered out with the authentic mask.
}
\label{fig:pipeline}
\vspace{-1cm}
\end{center}
\end{figure*}

%% file: tex/method.tex
\section{Definition of 3D GAN Inversion}

Similar to 2D GAN inversion, 3D GAN inversion aims to project an input image $I$ onto the manifold of a pre-trained unconditional 3D GAN model $G_{\text{3D}}(\cdot; \theta)$ parameterized by weight $\theta$. 
After inversion, $G_{\text{3D}}$ can reconstruct the image faithfully given the corresponding camera pose, synthesize content-consistent images in novel views, and facilitate downstream tasks like face editing.
One formulation of the 3D GAN inversion problem is defined as follows:

\begin{equation}
    w^* = \mathop{\arg\max}\limits_{w} = \mathcal{L} (G_{3D}(w, \pi; \theta), I), \label{eq:opt_w}
\end{equation}
where $w$ is the latent representation in $\mathcal{W}^+$ space and $\pi$ is the corresponding camera matrix of input image. The loss function $\mathcal{L}(\cdot, \cdot)$ is usually defined as pixel-wise reconstruction loss or perceptual loss.
In our settings, camera matrix $\pi$ is known, which is extracted by a pre-trained detector~\cite{deng2019accurate3dmm}. 
This formulation cares about the $\mathcal{W}^+$ space. 
However, the inversion in the $\mathcal{W}^+$ space is always not enough to capture local facial details, resulting in inaccurate reconstruction. 
Following the recent optimization-based 2D GAN inversion method~\cite{roich2021PTI}, we perform the inversion in the extended latent space for more accurate reconstruction, \textit{i.e.,} the combination of the $\mathcal{W}^+$ space and the parameter space.
The formulation is defined as: 
\begin{equation}
    w^*, \theta^* = \mathop{\arg\max}\limits_{w, \theta} = \mathcal{L} (G_{3D}(w, \pi; \theta), I). \label{eq:opt_w_theta}
\end{equation}
Note that $w$ and $\theta$ are optimized alternatively, \textit{i.e.,} $w$ is optimized using Eq.~(\ref{eq:opt_w}) first and then $\theta$ is optimized with the fixed $w^*$.

\section{The Proposed Approach}
\label{sec:method}

Our goal is to reconstruct a human face through a pre-trained 3D GAN given a single monocular image.   
The reconstruction is supposed to preserve authentic appearance texture and geometry shape in novel views.
Due to the limited information about geometry and texture from a single image, overfitting a single view tends to be trapped in geometry collapse, get the blurry texture and miss details in unseen views, especially when reconstructing a side face under an extreme pose. 
To overcome the issue of lacking information about other views, we introduce facial symmetry prior to promote inversion.
We propose a two-stage inversion pipeline, \textit{i.e.,} \textit{inversion for rough geometry} and \textit{joint optimization of geometry and texture. }
In the first stage, we obtain a rough geometry by optimizing the latent code $w$ using the original and mirror images in Sec.~\ref{subsec:inversion}. 
In the second stage, we refine the geometry and texture by optimizing the parameter $\theta$ with the depth-guided 3D warping and a set of designed constraints in Sec~\ref{subsec:pseudo}. 
An overview of our method is shown in Fig.~\ref{fig:pipeline}.

\input{image_tex/img_warp}
\input{image_tex/img_mask}

\subsection{Inversion with Symmetry for Rough Geometry }
\label{subsec:inversion}
The purpose of this stage is to learn a rough geometry as a pivot for further tuning.
To compensate for the missing information of unseen views, we resort to facial symmetry prior, \textit{i.e.,} the left face is almost the same as the right one.  
We simply flip the input image $I_{s}$ horizontally to get the mirror image $I_{m}$ whose corresponding camera pose $\pi_{m}$ can be calculated by multiplying a fixed matrix by the camera extrinsic parameters of $\pi_s$. 
The intrinsic parameters are unchanged.
The mirror image serves as the pseudo-projected image under a novel view. 
%
%

Since human faces are not always perfectly symmetric, the mirror image is just an approximation under the novel view. 
There exists inconsistent content between the original image and the mirror one if they have an overlapping face region, \textit{i.e.,} different colors in the position, referred as conflict content.  
The inversion should depend more on the original image and take partial useful information from the mirror one.
Furthermore, we observe that a frontal face can provide more effective information than a side face.
A nearly frontal face provides plenty of facial information, and we should trust less on its mirror image to avoid conflict in the overlapping region.   
While a side face provides information for only half one face, it has only a small overlapping conflict region with its mirror image. Hence, we should trust more on the mirror image.
We exploit an adaptive weighting strategy for the importance of the mirror image according to its yaw angle $\alpha_{\text{yaw}}$. 
We use a Gaussian function with respect to $\alpha_{\text{yaw}}$ to approximate the importance of different views.
The weight $\lambda_{m}$ of the mirror image is defined as:
\begin{gather}
    \mathcal{E}(x) = \frac{1}{\sigma \sqrt{2 \pi}} e^{- \frac{(x - \mu)^2}{2\sigma^2} }, \\
    \lambda_{m} = \begin{cases} 
1 - \mathcal{E}(\alpha_{\text{yaw}}),  & \text{if } \mathcal{E}(\alpha_{\text{yaw}}) \leq k ; \\
0, & \text{if } \mathcal{E}(\alpha_{\text{yaw}}) > k; 
\end{cases}
\end{gather}
where $\sigma$, $\mu$ and $k$ are hyper-parameters.
As a nearly frontal mirror face can compensate for very limited extra information for the original image, its weight $\lambda_{m}$ is clamped to $0$.

To optimize the latent code in  $\mathcal{W}^+$ space, the Perceptual loss~\cite{zhang2018lpips} is used to minimize the distance between the generated results and the original and mirror images. 
Following~\cite{karras2020analyzing,roich2021PTI}, a noise regularization term $\mathcal{L}_{n}(n)$ is employed to prevent the noise vector from containing vital information.
The objective in this stage is defined as follows:
\begin{equation}
\begin{aligned}
    \mathcal{L}_{\text{inv}} &= \mathcal{L}_{\text{LPIPS}}(G_{\text{3D}}(w, \pi_s; \theta), I_s) + \\ 
    & \lambda_{m} \mathcal{L}_{\text{LPIPS}}(G_{\text{3D}}(w, \pi_{m}; \theta), {I}_{m}) + \lambda_{n} \mathcal{L}_{n}(n),
\end{aligned}
\end{equation}
where $n$ is the noise vector and $\lambda_{n}$ is a trade-off parameter.
The generator is kept frozen at this stage.
Visual illustrations in Fig.~\ref{fig:ablation_mirror_w} show that the geometry can be greatly improved with the facial symmetry prior.

\subsection{Joint Optimization of Geometry and Texture}
\label{subsec:pseudo}
Though we obtain the rough geometry via the optimization of $w$ in the first stage, there is a distinct gap between the texture of the rendered face and that of the original one, even under the same camera pose.  
The rendered face shares a similar face geometry with the original one, but it becomes a different identity. 
In this stage, we optimize the generator's parameters $\theta$ to bridge the texture gap for identity preservation and refine the rough geometry as well. 
We design a geometry regularization constraint to avoid the model degrading to generate flattened geometry. 
Moreover, we construct a set of pseudo images in different views to provide supervision via depth-guided 3D warping.


\vspace{2pt}
\noindent \textbf{Geometry Regularization.}
We observe that optimizing the generator without any constraint on the geometry will cause the deviation of the geometry from the rough one, resulting in a flattened geometry similar to the case of inversion with a single image. 
To avoid the geometry drift during overfitting the texture, we regularize the optimized density obtained from the 3D volume of 3D GAN to be similar to that from the rough volume obtained in the first stage.
Specifically, with the fixed $w$, we generate depth maps $D$ from 3D GAN under different sampled views and calculate $\mathcal{L}_2$ distance between them with the corresponding depth maps $D_0$ generated from the un-tuned generator in the first stage: 
\begin{equation}
    \mathcal{L}_{\text{depth}} = \sum_{i \in \mathbb{S}}\Vert D^{i} - D^{i}_0\Vert_{2},
\end{equation}
where $\mathbb{S}$ is the sampled camera pose set.

\vspace{2pt}
\noindent \textbf{Depth-guided 3D Warping for Pseudo Supervision.}
Optimizing the generator with only two images is still not enough to capture the facial details, resulting in blurry effects around facial components such as eyes (see Fig.~\ref{fig:ablation_module}).  
Hence, we propose to construct pseudo images of different views for extra supervision using the rough geometry and the original and mirror images. 
Specifically, given the original image (source view) and the rough geometry, we can synthesize an image under a novel view (target view) by warping with 3D guidance. 
A coordinate pixel $p_{t}$ of the synthesized image in the target view can be obtained by projecting back onto the source view with the relative camera pose $\pi_{t \to s}$ and the camera intrinsic parameters $K$:
\begin{equation}
    p_{t \to s} = K \pi_{t \to s} D_t (p_t) K^{-1} p_t , 
\end{equation}
where $D_t(\cdot)$ is the depth map of the target view.
Since the projected coordinate $p_{t \to s}$ are continuous values, we can extract the color values from the original image with a differentiable bilinear sampling mechanism, \textit{i.e.}, $I_{s \to t} = I_s(p_{t \to s})$.
The low-resolution depth map will be upsampled to match the dimension of the image.

 \textit{Authentic Mask.}
Without distinguishing the foreground pixels from the background, the background pixels in the original image may be projected onto the foreground plane, leading to erroneous results.
To overcome this issue, we form a mask to indicate the visibility of pixels to filter invisible areas using the rendered depth values.
Specifically, we can get the projected depth value $D_s(p_{t \to s})$ via sampling from the depth map in the source view.
Here we employ the euclidean distance between $D_s(p_{t \to s})$ and the depth map $D_t(p_{t})$ in the target view to calculate the mask. 
A large distance indicates the pixel $p_t$ is invisible. 
To ensure the projected pixels are located on the front visible surface, we only preserve the area where the distance is under a threshold $\tau$:
\begin{equation}
    M(p_{t}) = \| D_{t}(p_t) - D_{s}(p_{t \to s}) \| < \tau.
\end{equation}
Furthermore, due to the poor depth estimation of the background, only the facial part would be warped.
We warp the facial mask of the source view to the target view and multiply it with the visibility mask $M(p_{t})$ to get the authentic mask $M_{t}$. An example is shown in Fig.~\ref{fig:mask}.
After multiplying the mask $M_{t}$ with the warped image $I_{s \to t}$, the resulting image can be used for supervision.

\vspace{2pt}
\noindent \textbf{Adjacent View Warping.}
Fig.~\ref{fig:warp} illustrates the warping results of two examples. 
When the yaw angle between the source and target views increases, the warping results have more distortions and become less authentic.
Therefore, it is intuitive to abandon the pseudo images of the target views that deviate a lot from the source view.
Empirically, a frontal face can be warped by a wider range of yaw angles than a side face to get authentic pseudo images.
The variance of sampling yaw angles for constructing pseudo images is set to a fixed ratio of $\lambda_{m}$ that depends on the viewpoint mentioned in Sec.~\ref{subsec:inversion}.
The LPIPS loss~\cite{johnson2016perceptual_loss} is used to compute the multi-view pixel-wise distance as follows: 
\begin{equation}
    \mathcal{L}_{\text{adj}} =  \mathcal{L}_{\text{LPIPS}}(M_t \cdot G_{\text{3D}}(w, \pi_{t}; \theta), M_t \cdot I_{s \to t}).
\end{equation}

Although the pseudo images of several unseen adjacent views around the source view have been constructed, it brings marginal improvements on remote views.
Especially for a side face, the pseudo images of the remote views are blurry and have incomplete texture (see Fig.~\ref{fig:warp}).
Therefore, we also construct pseudo images of the adjacent views around the view of the mirror image. 

Since the conflict region between the original and mirror images has a side effect on the generator optimization process, resulting in blurry effects on rendered images, even reconstructing the source view (see Fig.~\ref{fig:ablation_mirror_g}), 
we propose to take partial meaningful information from the symmetric views without harming the original inversion quality. 
We compute the similarities only for facial components, rather than the whole face region. 
Besides, instead of using a pixel-wise loss, we exploit the contextual loss~\cite{mechrez2018cxloss} to improve the texture quality.
The loss for symmetric views is defined as:
\begin{equation}
    \mathcal{L}_{\text{sym}} = \sum_{\text{c} \in \mathbb{F}} \mathcal{L}_{\text{CX}}(\text{ROI}^c(G_{\text{3D}}(w, \pi_{t}; \theta)), \text{ROI}^c(I_{m \to t})), 
\end{equation}
where $I_{m \to t}$ is the pseudo image of the viewpoint $\pi_t$ warped from the mirror image $I_m$. 
$\text{ROI}^c(\cdot)$ refers to the region of interest component $c$ from the collection $\mathbb{F} = \{\text{eyes}, \text{nose}, \text{mouth}\}$.

The reconstruction loss between the original image and its corresponding rendered image is still in use to ensure the quality of the initial perspective, which is defined as:
\begin{equation}
    \mathcal{L}_{\text{ori}} = 
    \mathcal{L}_{2}(G_{\text{3D}}(w, \pi_{s}; \theta), I_{s}) + \mathcal{L}_{\text{LPIPS}}(G_{\text{3D}}(w, \pi_{s}; \theta), I_{s}).
\end{equation}
The overall objective of optimizing the generator's parameters is defined as:
\begin{equation}
    \mathcal{L}_{\text{opt}} =  \mathcal{L}_{\text{ori}} +
    \lambda_{\text{adj}} \mathcal{L}_{\text{adj}} +
    \lambda_{\text{sym}} \mathcal{L}_{\text{sym}} +
    \lambda_{\text{depth}} \mathcal{L}_{\text{depth}}.
\end{equation}
The trade-off hyper-parameters are set as follows: $\lambda_{\text{adj}}=0.1$, $\lambda_{\text{sym}}=0.05$, and $\lambda_{\text{depth}}=1$.


%% file: image_tex/img_warp.tex
\begin{figure*}[t]
\begin{center}
\centerline{\includegraphics[width=1\linewidth]{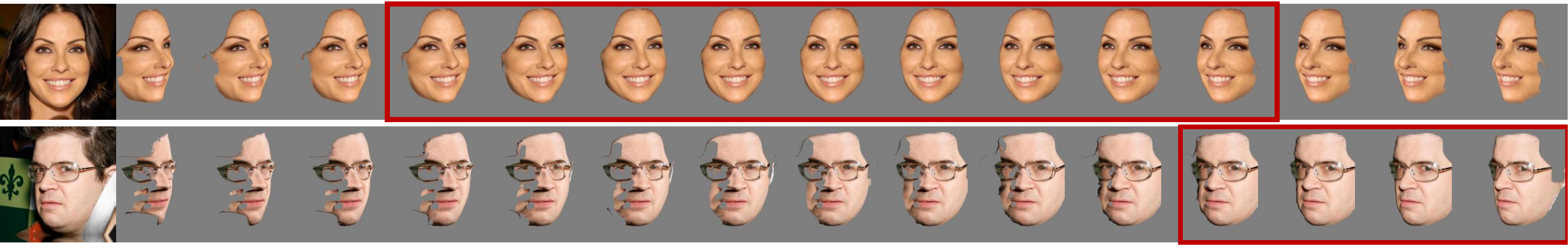}}
\caption{
Visualization of warped pseudos. 
The red bounding box contains the range of employed pseudos, depending on the yaw angle of the input image.
A frontal face can be warped by a wider range of yaw angles than a side face to get authentic pseudos.
}
\label{fig:warp}
\vspace{-1cm}
\end{center}
\end{figure*}

%% file: image_tex/img_mask.tex
\begin{figure}[t]
\begin{center}
\centerline{\includegraphics[width=1\linewidth]{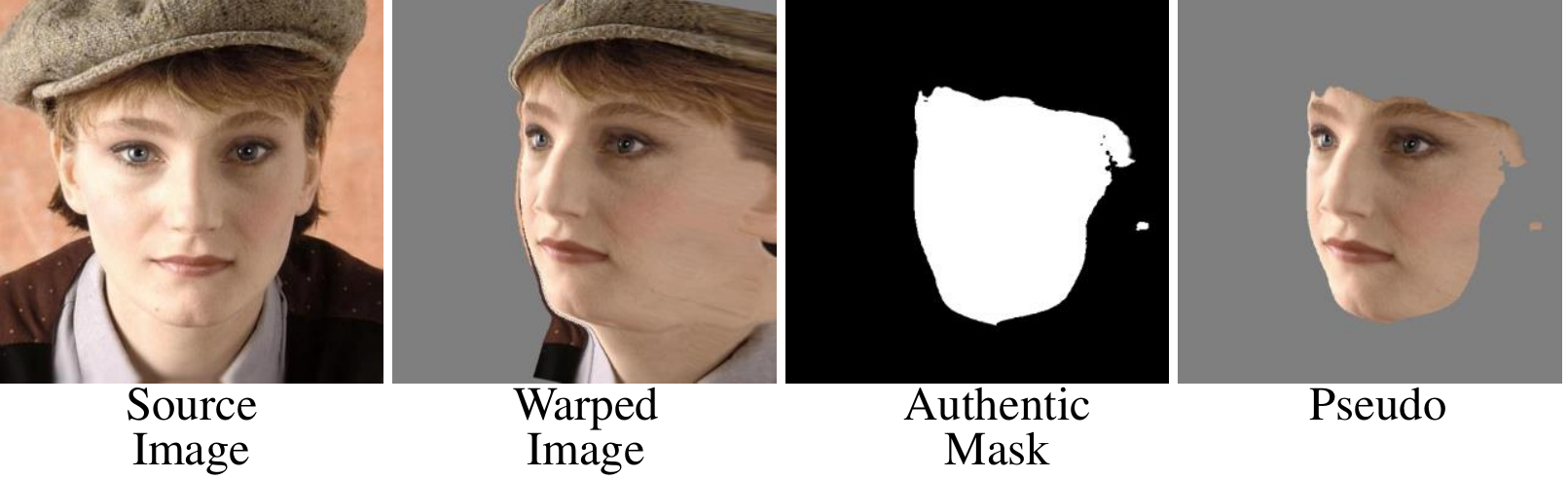}}
\caption{
Visualization of authentic mask and warped pseudo.
}
\label{fig:mask}
\vspace{-1cm}
\end{center}
\end{figure}

%% file: tex/experiment.tex
\input{image_tex/img_quality}

\section{Experiments}
\label{sec:experiment}

\subsection{Experimental Settings}

\noindent \textbf{Datasets.}
We conduct the experiments on human faces datasets.
For all experiments, we select EG3D~\cite{chan2022eg3d} as our 3D GAN prior, which is pre-trained on FFHQ dataset~\cite{karras2019ffhq}.
We verified quantitative metrics on CelebA-HQ test dataset~\cite{lee2020celeba}.
We further evaluated on MEAD~\cite{kaisiyuan2020mead}, a multi-view high-quality video dataset.
The first frame from each viewpoint video of $10$ identities is extracted for testing.

\vspace{2pt}
\noindent \textbf{Metrics.}
We evaluate image reconstruction quality and similarity with the following metrics: mean squared error~(MSE), 
perceptual similarity loss~(LPIPS)~\cite{zhang2018lpips}, structural similarity~(MS-SSIM), and identity similarity~(ID) by employing a pre-trained face recognition network~\cite{deng2019arcface}.

\vspace{2pt}
\noindent \textbf{Baselines.}
We mainly compare our methods with optimization-based 2D GAN inversion methods.
SG2~\cite{stylegan2} directly inverts real images into $\mathcal{W}$ space with an optimization scheme. \cite{abdal2019image2stylegan} extends the inversion into $\mathcal{W}^+$ space, denoted by SG2 $\mathcal{W}^+$.
PTI~\cite{roich2021PTI} would further tune generator parameters in a second stage. 
For a fair comparison, both PTI and ours first optimize the latent for $500$ steps and then fine-tune the generator for $1,000$ steps, while SG2 and SG2 $\mathcal{W}^+$ optimize the latent for $1,500$ steps.

\subsection{Reconstruction and Novel View Synthesis}

\input{table/celebahq}

\input{image_tex/img_metric}

\noindent \textbf{Qualitative Evaluation.}
Fig.~\ref{fig:quality} presents a qualitative comparison of texture and geometry quality of different views. 
As for the original view, our method is able to inverse challenging details such as earrings, make-up, and wrinkles, which demonstrates that we do not sacrifice the original reconstruction performance.
When the camera rotates to novel views, images generated from 2D inversion methods present a twisted appearance, due to the nearly flattened geometry shape.
Since SG2 does not deviate too far from the initial GAN space, it can generate a portrait with a structured geometry, but fails to preserve the identity.
Our method is capable of maintaining authentic and consistent geometry in novel views along with a sharp appearance, even when rotated to an extreme pose.

\vspace{2pt}
\noindent \textbf{Quantitative Evaluation.} 
The reconstruction metrics of the original view are shown in Table~\ref{tb:criterion}.
As can be seen, the results align with our qualitative evaluation as we achieved comparable scores to the current 2D \textit{state-of-the-art} inversion methods~\cite{roich2021PTI}.
The MSE, LPIPS, and ID similarities of ours are further improved, which can be attributed to the employment of $\mathcal{W}^+$ latent space.
Following EG3D, we evaluate shape quality by calculating $\mathcal{L}_{2}$ for pseudo-ground-truth depth-maps (\textit{Depth}) generated from DECA~\cite{feng2021deca}, and poses (\textit{Pose}) estimated from synthesized images.

We also use identity similarity to evaluate the identity preservation of the synthesized novel views. 
Given a portrait, we synthesize a novel view image under the symmetric camera pose of the portrait. 
The similarity between the synthesized image and the flipped image portrait is calculated.
The results are shown in Fig.~\ref{fig:metric}. 
It can be observed that when the yaw angle of a portrait is small, all methods can perform well with a high similarity score. 
But when the yaw angle is large, only our method can maintain a high score, while other methods encounter a sharp performance drop due to the inaccurate geometry.   
As we employ the symmetry prior and the adjacent pseudo supervision, the rendered faces can better preserve the texture and geometry. 
These results demonstrate that we can achieve an identity-consistent 3D inversion.

\input{image_tex/img_mead}
\input{table/mead}

\vspace{2pt}
\noindent \textbf{Evaluation on MEAD.}
To get a comprehensive understanding of the performance of our method, we evaluate on MEAD, a multi-view dataset.
The quantitative comparison between the reconstruction portraits and the ground truth in different views is shown in Tab.~\ref{tb:mead_criterion}.
PTI~\cite{roich2021PTI} and our method achieve comparable performance when given a frontal portrait.
When the view of the input face has an offset from the canonical one
, our method surpasses PTI distinctly.
Our metrics remain stable as the yaw angle becomes larger while the performance of PTI degrades significantly.
The qualitative results are shown in Fig.~\ref{fig:mead}.
The geometry shape of PTI suffers from the flattening phenomenon.
In contrast, our method can generate a consistent geometry and texture in novel views.


\input{image_tex/img_ablation_mirror_w}
\input{image_tex/img_ablation_mirror_g}

\subsection{Evaluation of Symmetry Prior}
To understand the importance of the \textit{symmetry prior},
we perform an ablation study by conducting the inversion with or without using the prior. 
The visual results are shown in Fig.~\ref{fig:ablation_mirror_w}.
%
Both approaches can obtain good geometries in the original view.
However, in the first row, the geometry of the woman with a thin face turns to be obese as the camera gradually rotates, which aligns with its rendered image.
The second row shows that the geometry and the rendered image maintain a better view consistency.
We even find that, with the auxiliary view, some expression details can be strengthened, such as the slightly opened mouth.

The symmetry prior cannot be directly employed in the optimization stage because there exist asymmetric areas in a human face. 
Optimizing the conflict areas will lead to poor results.
As shown in Fig.~\ref{fig:ablation_mirror_g}, the slanted hair and the single earring in the source image mismatch those in the mirror one.
In the first row, when simply using both two images to optimize the generator, 
the reconstruction quality suffers from degradation.
Novel views synthesized by the vanilla version will encounter incorrect texture and blurry results in the conflict areas.
Our method can handle such asymmetric cases without the quality worsening by filtering out conflict areas with the designed constraints.
Hair, teeth, and other details are consistent in different views, which validates the effectiveness of the proposed constraints.

\input{image_tex/img_editing}

\subsection{View-consistent Face Editing}
Editing a facial image should preserve the original identity while performing a meaningful and visually plausible modification. 
We extend our methods to downstream editing tasks to validate that the 3D GAN inversion process does not degrade the editability of the original generator.
We follow StyleCLIP~\cite{patashnik2021styleclip} to achieve text-guided semantic editing and StyleGAN-NADA~\cite{gal2022stylegannada} for stylization, shown in Fig.~\ref{fig:editing}.
The editing operation not only influences the original view but also changes the novel view's appearance consistently.
It demonstrates that our inversion solution retains the properties in the original space of the generator and can be associated with other editing methods flexibly.

\input{image_tex/img_ablation_module}

\subsection{Ablation Study}

\vspace{2pt}
\noindent \textbf{Adjacent Warping.}
Recall that we employ depth-guided warping to create pseudo supervision to improve the texture quality of novel views.
In Fig.~\ref{fig:ablation_module},
we can find that this operation can enhance facial component details such as eyelashes and teeth, improving the overall visual quality.

\vspace{2pt}
\noindent \textbf{Depth Regularization.}
Since supervision signals all come from RGB images, there is no explicit geometry supervision to ensure shape correctness. 
The shape is prone to drift to overfit the single image. 
Unnatural distortions will appear in novel views with the drifted shape. 
In the third column of Fig.~\ref{fig:ablation_module}, the jaw and nose are elongated with no constraints. 
With depth regularization, geometry will be calibrated within reasonable limits.

\vspace{2pt}
\noindent \textbf{Two-stage Optimization.}
The joint optimization stage via utilizing a large parameter space can further improve texture, allowing to reconstruct the out-of-domain details, \textit{e.g.,} auspicious mole,
as shown in the last column of Fig.~\ref{fig:ablation_module}.



%% file: image_tex/img_quality.tex
\begin{figure*}[t]
\begin{center}
\centerline{\includegraphics[width=1\linewidth]{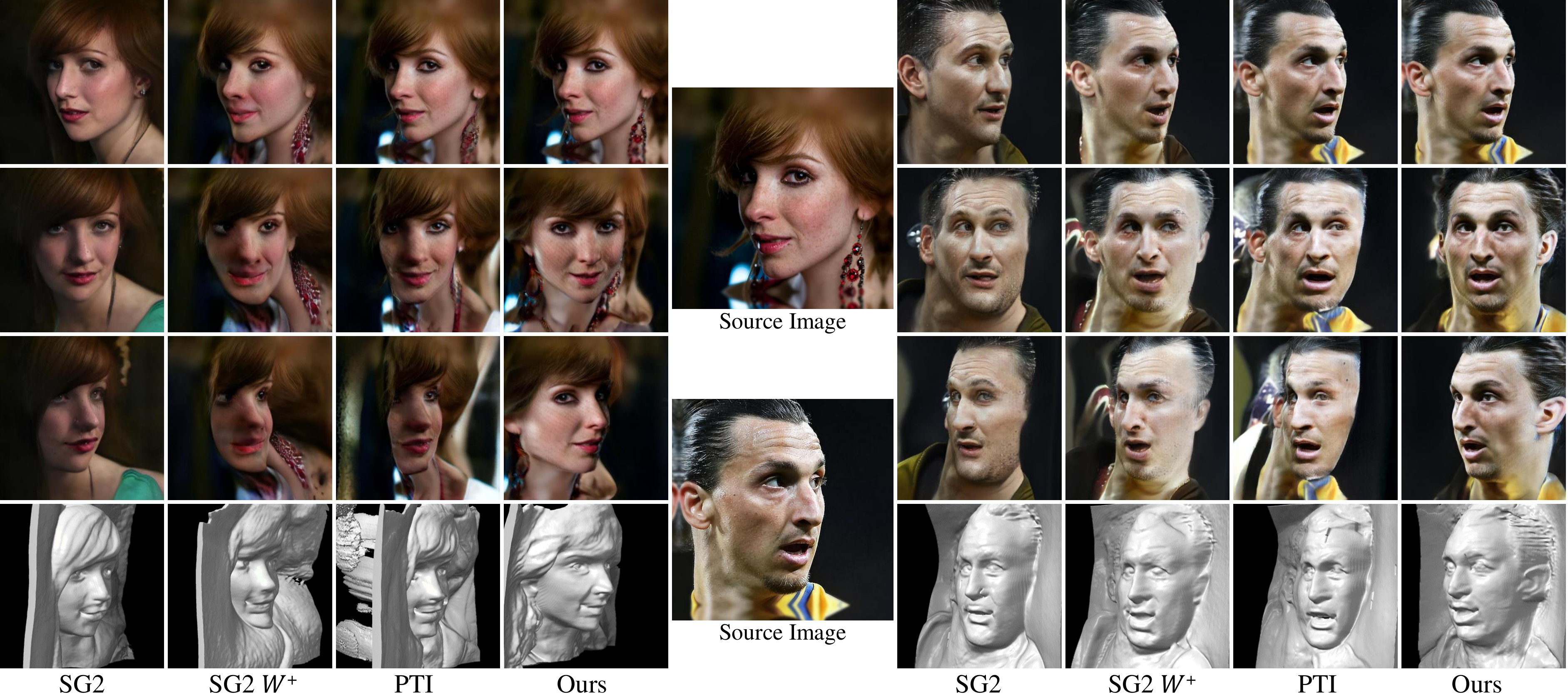}}
\caption{
Qualitative comparisons with \textit{state-of-the-art} methods on novel view synthesis.
The reconstruction quality of the original view is presented in the first row.
The texture and geometry in novel views are shown in the rest rows.
}
\label{fig:quality}
\vspace{-1cm}
\end{center}
\end{figure*}

%% file: table/celebahq.tex
\begin{table}[t]
\scriptsize
\setlength\tabcolsep{4pt}
\renewcommand\arraystretch{1.1}
\begin{tabular}{c|cccccc}
\toprule
Method & MSE $\downarrow$ & LPIPS $\downarrow$ & MS-SSIM $\downarrow$ & ID $\uparrow$ & Pose $\downarrow$ & Depth $\downarrow$ \\ 
\hline
SG2~\cite{stylegan2} & 0.0881 & 0.3231 & 0.3557 & 0.8209 & 0.043 & 0.0505 \\
SG2 $\mathcal{W}^+$~\cite{abdal2019image2stylegan} & 0.0439 & 0.2261 & 0.2483 & 0.8735 & 0.040 & 0.0500 \\
PTI~\cite{roich2021PTI} & 0.0084 & 0.0920 & \textbf{0.0980} & 0.9432 & 0.037 & 0.0510 \\
SPI~(Ours) & \textbf{0.0082} & \textbf{0.0865} & 0.0991 & \textbf{0.9470} & \textbf{0.036} & \textbf{0.0476} \\
\hline
\end{tabular}
\caption{Quantitative comparison on CelebA-HQ~\cite{lee2020celeba}.}
\label{tb:criterion}
\vspace{-9pt}
\end{table}

%% file: image_tex/img_metric.tex
\begin{figure}[t]
\begin{center}
\centerline{\includegraphics[width=1\linewidth]{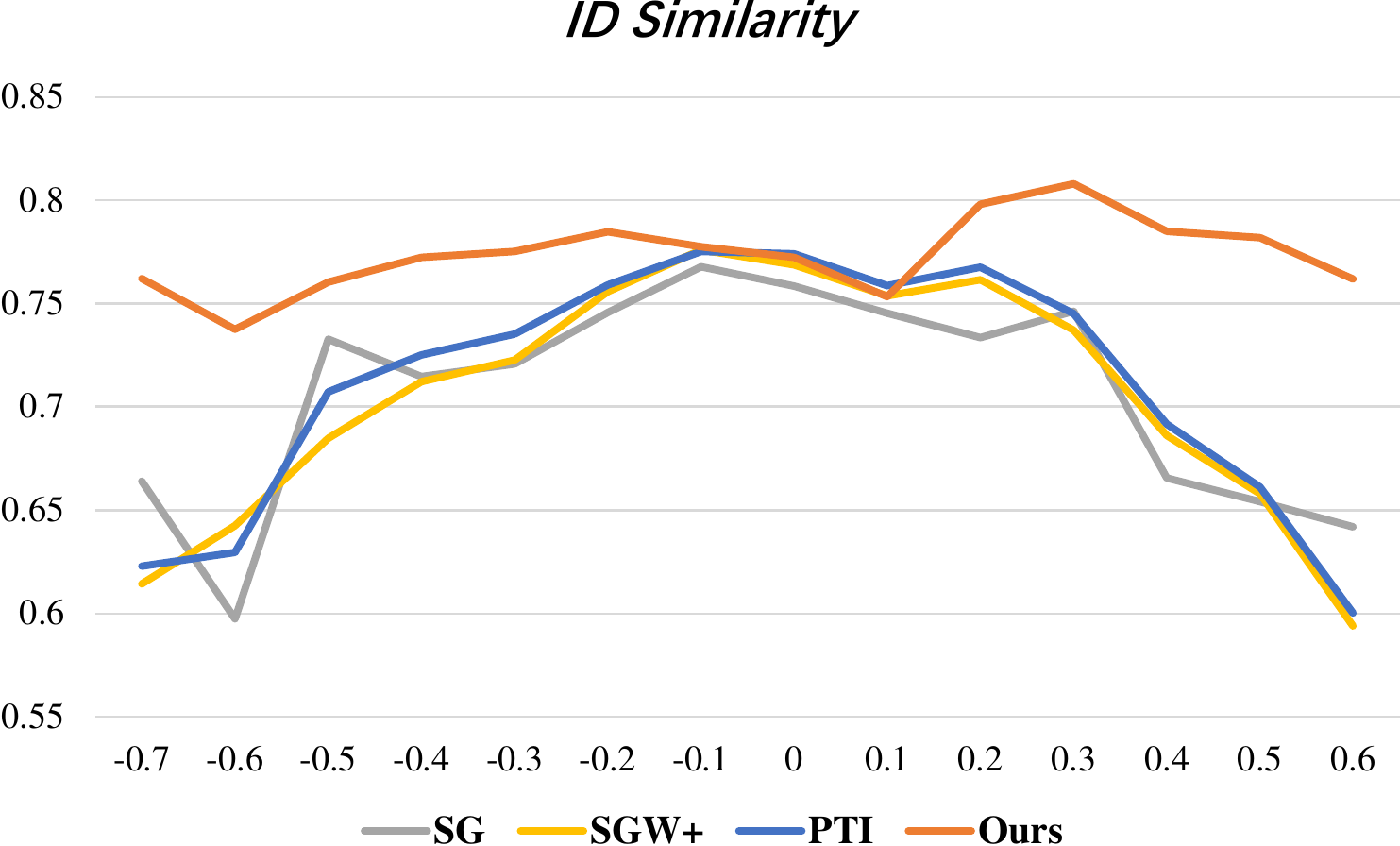}}
\caption{
Comparison of identity preservation in novel views. The x-axis represents the yaw angle of the input image. `0' indicates the frontal face. 
}
\label{fig:metric}
\vspace{-1cm}
\end{center}
\end{figure}

%% file: image_tex/img_mead.tex
\begin{figure}[t]
\begin{center}
\centerline{\includegraphics[width=1\linewidth]{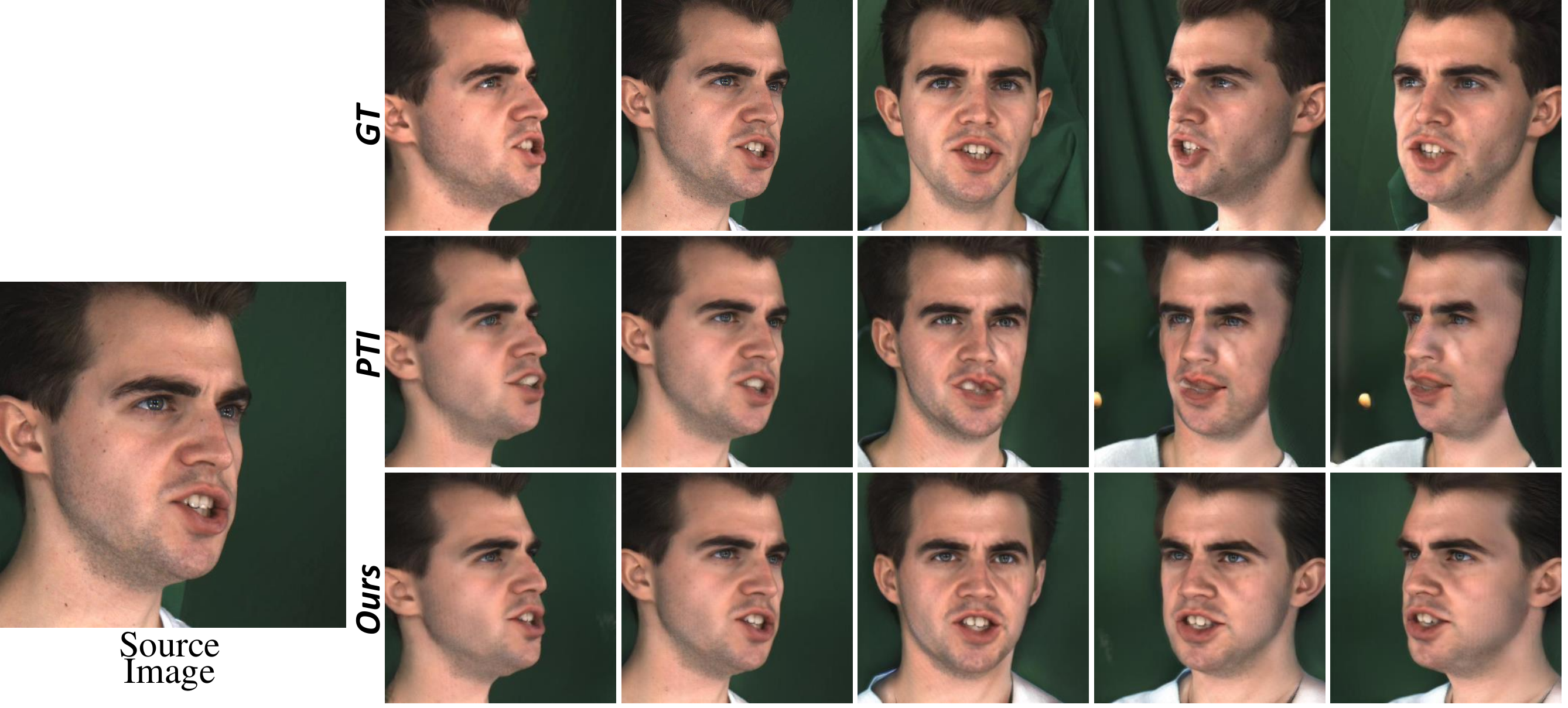}}
\caption{
Qualitative comparisons with PTI~\cite{roich2021PTI} on MEAD~\cite{kaisiyuan2020mead}.
}
\label{fig:mead}
\vspace{-1cm}
\end{center}
\end{figure}

%% file: table/mead.tex
\begin{table}[t]
\setlength\tabcolsep{4pt}
\renewcommand\arraystretch{1.1}
\begin{tabular}{c|ccccc}
\toprule
Method & View & MSE $\downarrow$ & LPIPS $\downarrow$ & MS-SSIM $\downarrow$ & ID $\uparrow$ \\ 
\hline
PTI & \multirow{2}*{F} & \textbf{0.03204} & \textbf{0.2971} & \textbf{0.2070} & \textbf{0.8445} \\
Ours & ~ & 0.03296 & 0.3088 & 0.2135 & 0.8388 \\
\hline
PTI & \multirow{2}*{L30} & 0.04355 & 0.2992 & 0.2274 & 0.8446 \\
Ours & ~ & \textbf{0.03399} & \textbf{0.2796} & \textbf{0.2025} & \textbf{0.8469} \\
\hline
PTI & \multirow{2}*{L60} & 0.08255 & 0.3902 & 0.3143 & 0.7568 \\
Ours & ~ & \textbf{0.04069} & \textbf{0.3113} & \textbf{0.2379} & \textbf{0.8272} \\
\hline
PTI & \multirow{2}*{R30} & 0.04574 & 0.3110 & 0.2393 & 0.8383 \\
Ours & ~ & \textbf{0.03203} & \textbf{0.2807} & \textbf{0.2057} & \textbf{0.8529} \\
\hline
PTI & \multirow{2}*{R60} & 0.07865 & 0.3829 & 0.3106 & 0.7995 \\
Ours & ~ & \textbf{0.04541} & \textbf{0.3160} & \textbf{0.2400} & \textbf{0.8335} \\
\hline
\end{tabular}
\caption{Quantitative comparison on MEAD~\cite{kaisiyuan2020mead}.
View denotes the yaw angle of the input image. F is frontal, L is left side, and R is right side. 30 and 60 are the rotation degrees.
Each time we use one view as the inversion input and use all 5 views as ground truth for evaluation.
The average performance of 4 unseen views and 1 seen view is reported. 
}
\label{tb:mead_criterion}
\vspace{-2pt}
\end{table}

%% file: image_tex/img_ablation_mirror_w.tex
\begin{figure}[t]
\begin{center}
\centerline{\includegraphics[width=1\linewidth]{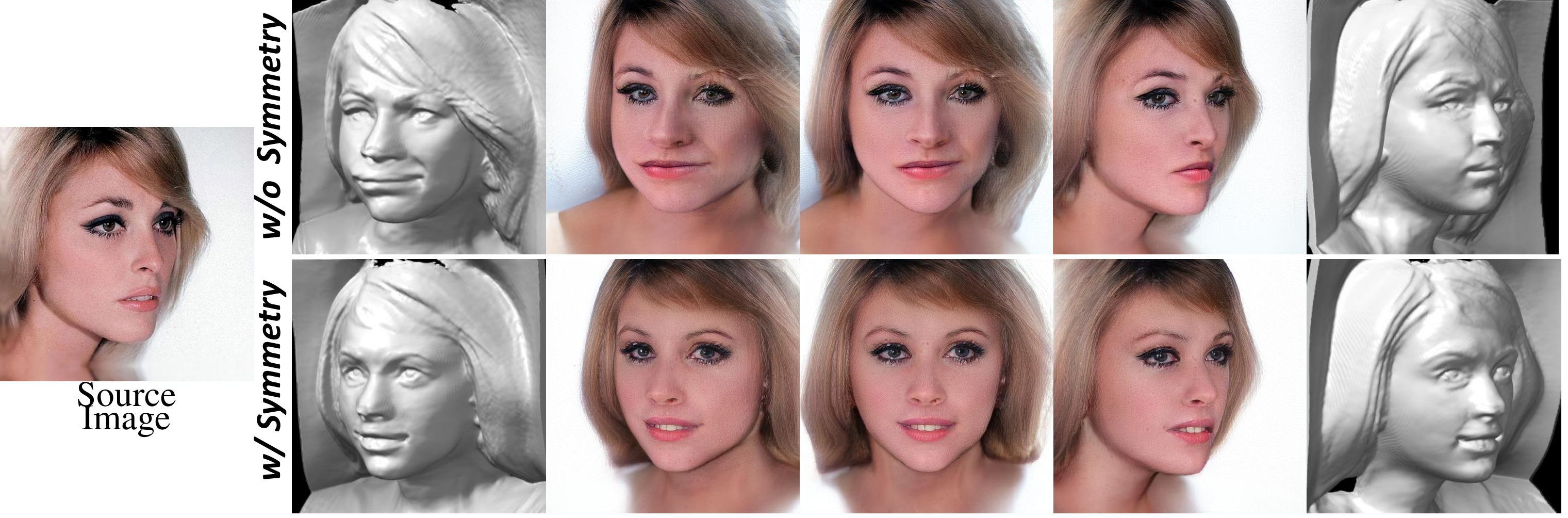}}
\caption{
Ablation study of facial symmetry prior.
}
\label{fig:ablation_mirror_w}
\vspace{-1.1cm}
\end{center}
\end{figure}

%% file: image_tex/img_ablation_mirror_g.tex
\begin{figure}[t]
\begin{center}
\centerline{\includegraphics[width=1\linewidth]{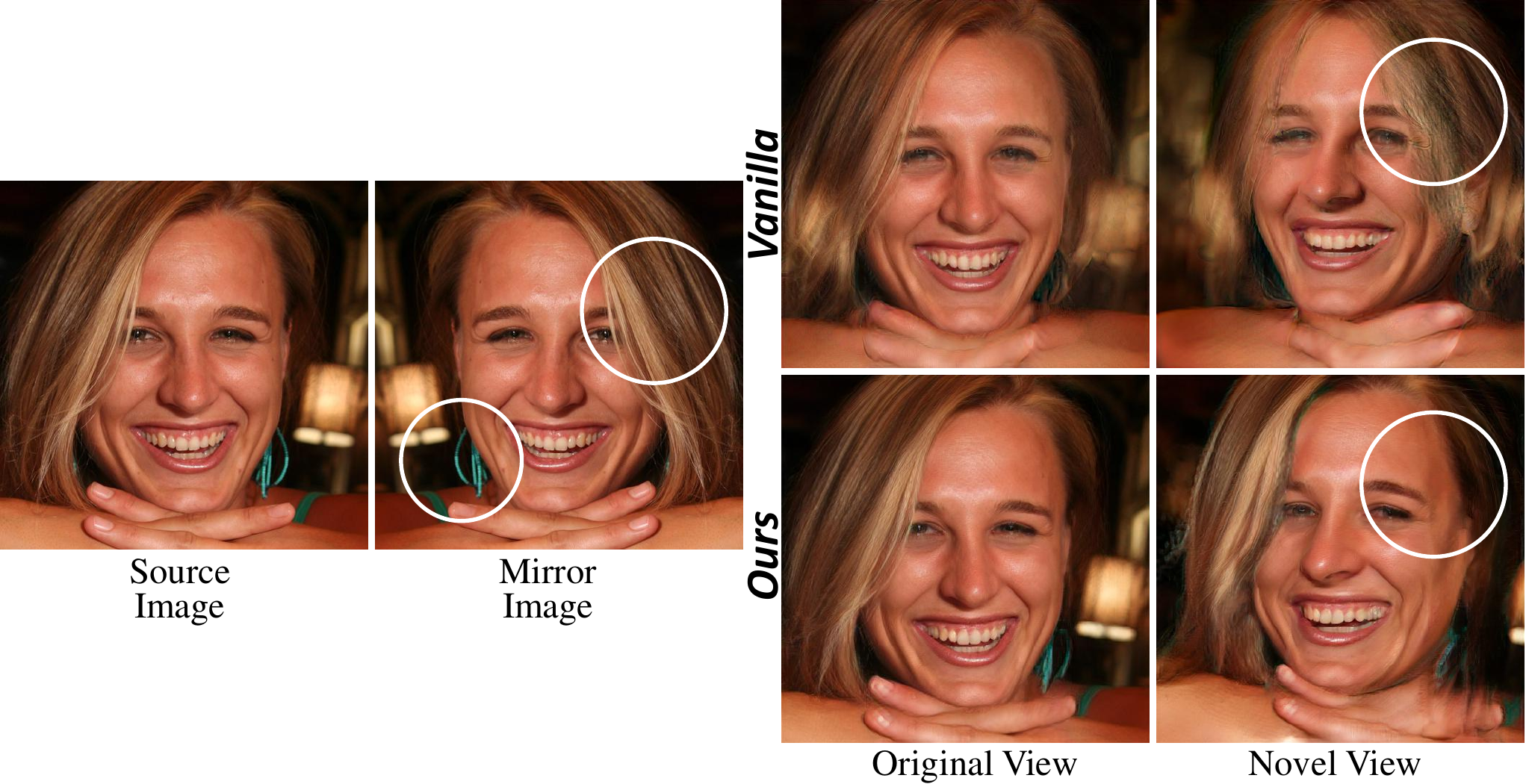}}
\caption{
Ablation study of authentic mask. \textit{Vanilla} denotes simply using the full mirror image for supervision. While \textit{Ours} filters out conflict areas with the designed constraints.
}
\label{fig:ablation_mirror_g}
\vspace{-1cm}
\end{center}
\end{figure}

%% file: image_tex/img_editing.tex
\begin{figure}[t]
\begin{center}
\centerline{\includegraphics[width=1\linewidth]{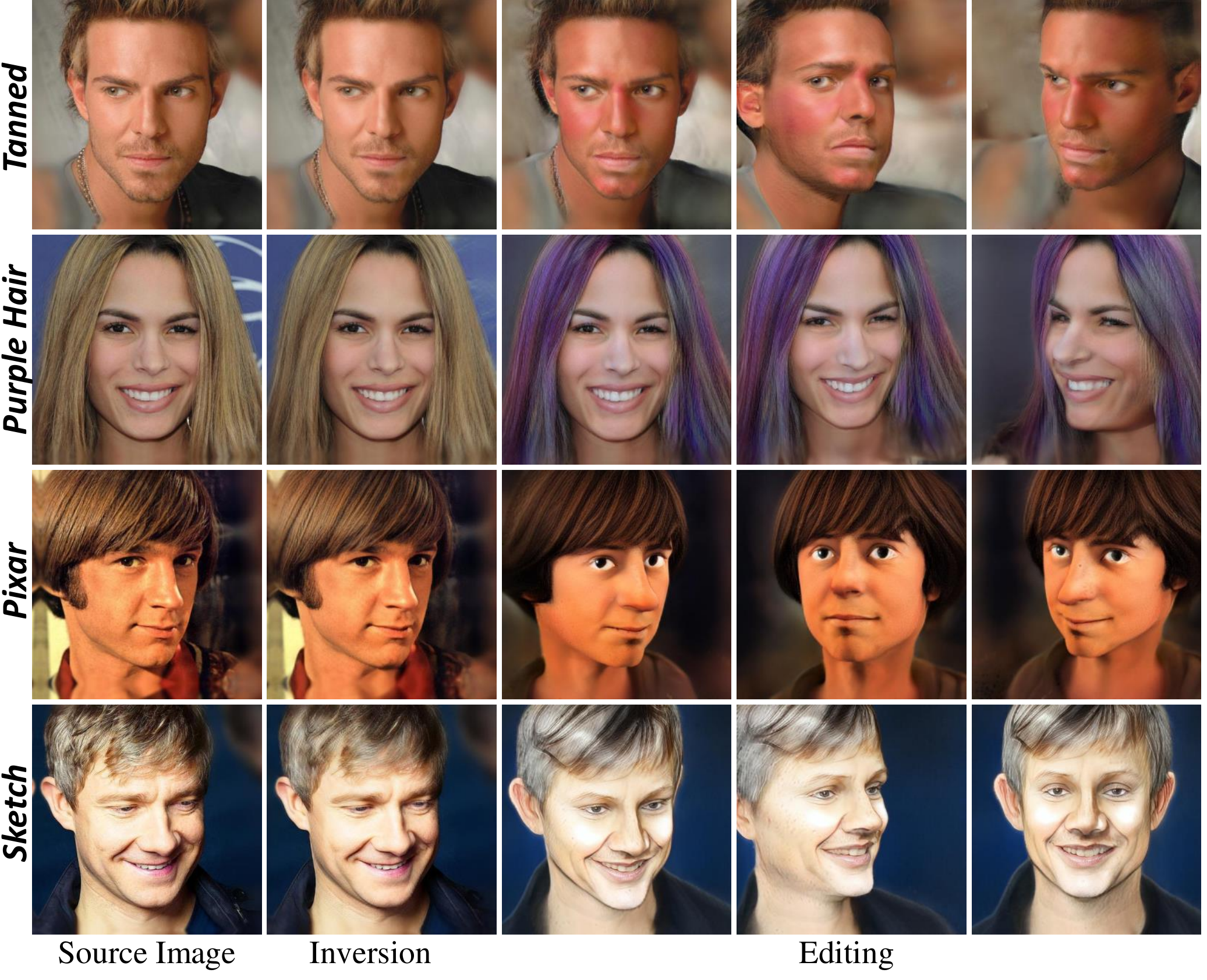}}
\caption{
Editing results incorporated with \cite{patashnik2021styleclip} and \cite{gal2022stylegannada}.
}
\label{fig:editing}
\vspace{-1cm}
\end{center}
\end{figure}

%% file: image_tex/img_ablation_module.tex
\begin{figure}[t]
\begin{center}
\centerline{\includegraphics[width=1\linewidth]{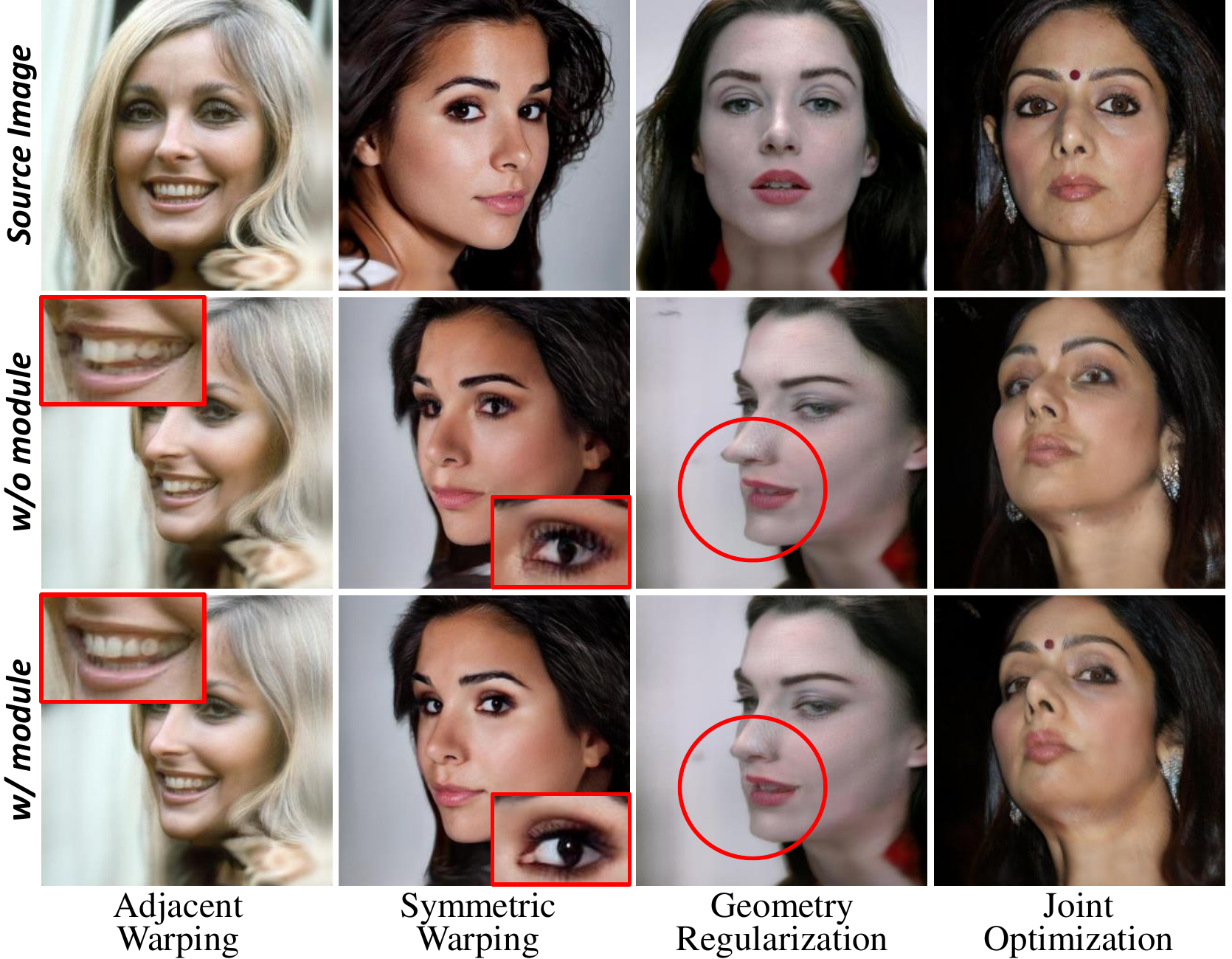}}
\caption{
Ablation study of different designed modules.
}
\label{fig:ablation_module}
\vspace{-1cm}
\end{center}
\end{figure}

%% file: tex/conclusion.tex
\section{Conclusion}
\label{sec:conclusion}
We propose a novel 3D GAN inversion method with facial symmetry prior.
As demonstrated in massive experiments, our method can support 3D reconstruction at extreme angles with robust geometry. 
With the designed constraints on texture and geometry, the reconstructed portraits are high-fidelity and possess consistent identity across different views. 
Besides, the proposed method enables various downstream applications without compromising faithfulness and photorealism.

\vspace{2pt}
\noindent \textbf{Limitation and Future Works.}
Since the effect of illumination is ignored in our assumption, the illumination is modeled implicitly.
During the fitting process of the given image with symmetry prior, light sources sometimes become perfectly symmetrical and distorted.
We will attempt to settle the problem via modeling illumination explicitly with albedo and normal in future work.

\vspace{2pt}
\noindent\textbf{Acknowledgement.}
This work was partly supported by the National Natural Science Foundation of China (Grant No. U1903213) and the Shenzhen Science and Technology Program (JCYJ20220818101014030, ZDSYS20200811142605016).
%
This work was partly supported by a UKRI Future Leaders Fellowship [grant number G104084].

%% file: tex/appendix.tex
\newpage
\appendix
\section*{Appendix}




\section{Ablation Study on MEAD}
To further verify the designed modules and strategy of our method, we conduct ablation study experiments on a multi-view dataset, MEAD~\cite{kaisiyuan2020mead}.
The quantitative results are shown in Tab.~\ref{tb:mead_ablation_module}.
`$\mathcal{W}^+$ Inversion' denotes optimizing latent in $\mathcal{W}^+$ space with $500$ iterations using only the ground truth image.
`+ Symmetry Prior' denotes optimizing latent in $\mathcal{W}^+$ space with $500$ iterations employing both original and symmetric view.
`+ Joint Optimization' would further optimize generator parameters with $1,000$ iterations.
`+ Geometry Regularization' would regularize the  shape correctness during the joint optimization process. 
`+ Warping Pseudo' would introduce depth-guided 3D  warping pseudos for supervision.
It can be seen that the symmetry prior can strongly boost the vanilla inversion method, especially when inputting a side face~(\textit{e.g.}, L60, R60).
The other designed modules enhance the baselines to a different extent from the rest columns.
The results demonstrate that reasonable geometry of our method can help the model synthesize robust and consistent texture, which aligns with the conclusion of the manuscript.

\section{Comparison with 2D GAN Inversion.}
We apply rotation editing~\cite{shen2020interfacegan} directions to the latent code of StyleGAN-2~\cite{stylegan2} to mimic the camera rotation.
The comparison is shown in Fig.~\ref{fig:ablation_2d_3d}.
The rotation of~\cite{shen2020interfacegan}  makes changes to the identity.
While our method can generate consistent and high-fidelity portraits in different views.

\section{Additional Results on In-the-wild Images}
Following the baseline comparison in the manuscript, we provide additional inversion results on in-the-wild images shown in Fig.~\ref{fig:appendix_1} and Fig.~\ref{fig:appendix_2}, which demonstrate the effectiveness of our 3D GAN inversion method.

\newpage
\input{image_tex/img_ablation_2d_3d}
\input{table/mead_ablation.tex}

\input{image_tex/img_ablation_appendix.tex}


%% file: image_tex/img_ablation_2d_3d.tex
\begin{figure}[h]
\begin{center}
\centerline{\includegraphics[width=1\linewidth]{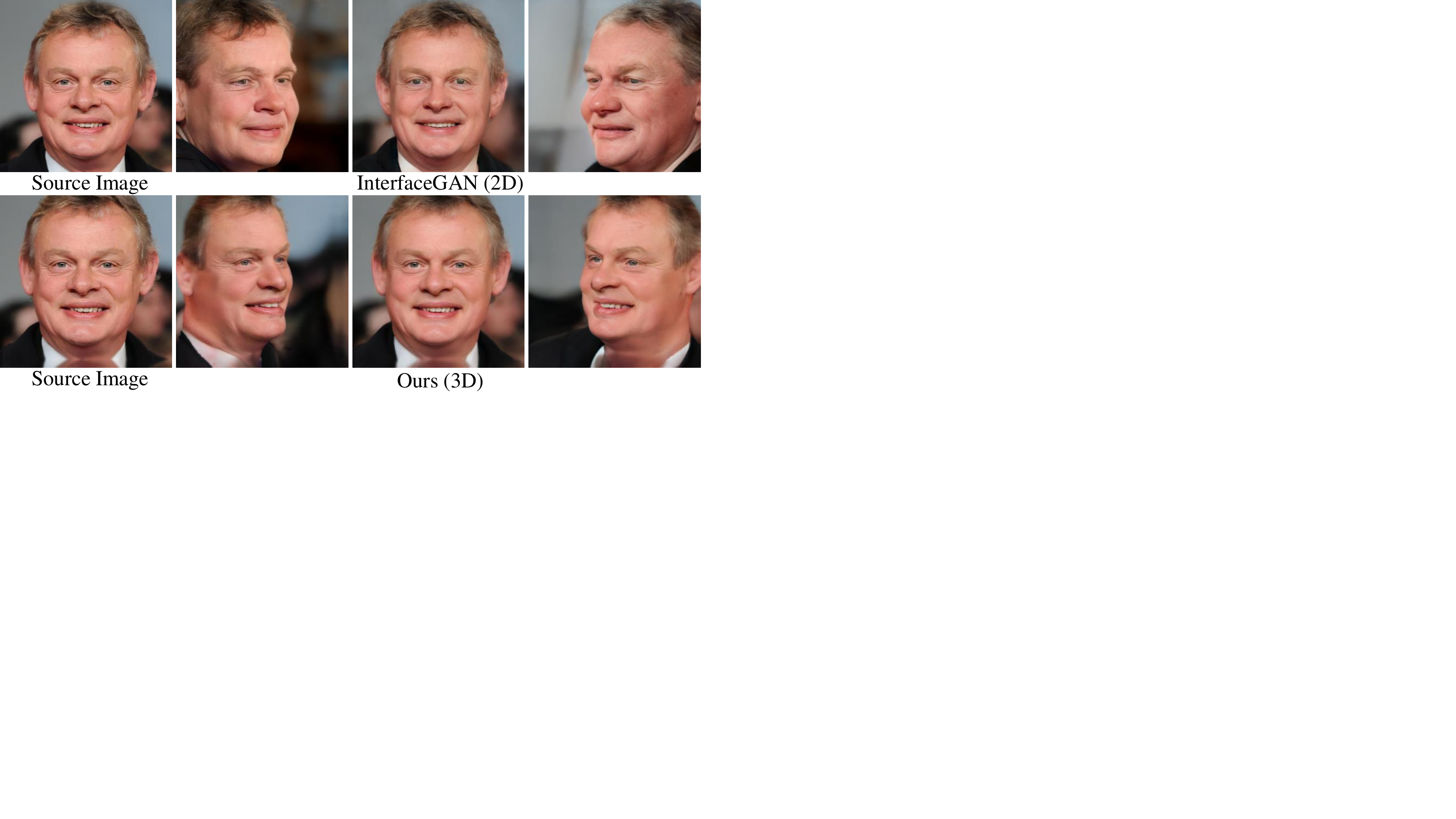}}
\caption{
Comparison of 2D and 3D GAN inversion along with viewpoint change.
}
\label{fig:ablation_2d_3d}
\vspace{-1cm}
\end{center}
\end{figure}

%% file: table/mead_ablation.tex
\begin{table*}[t]
\begin{tabular*}{\textwidth}{c @{\extracolsep{\fill}} |ccccc}
\toprule
Method & View & MSE $\downarrow$ & LPIPS $\downarrow$ & MS-SSIM $\downarrow$ & ID $\uparrow$ \\ 
\hline
$\mathcal{W}^+$ Inversion  & \multirow{5}*{F} & 0.04853 & 0.3358 & 0.2681 & 0.8124 \\
+ Symmetry Prior & ~ & 0.04119 & 0.3253 & 0.2531 & 0.8148 \\
+ Joint Optimization  & ~ & 0.03474 & 0.3161 & 0.2210 & 0.8364 \\
+ Geometry Regularization \textcolor{white}{in} & ~ & 0.03315 & 0.3123 & 0.2158 & 0.8363 \\
+ Warping Pseudo~(Ours) & ~ & \textbf{0.03296} & \textbf{0.3088} & \textbf{0.2135} & \textbf{0.8388} \\

\hline

$\mathcal{W}^+$ Inversion & \multirow{5}*{L30} & 0.05158 & 0.3286 & 0.2659 & 0.8111 \\
+ Symmetry Prior & ~ & 0.04278 & 0.3002 & 0.2375 & 0.8245 \\
+ Joint Optimization  & ~ & 0.03321 & 0.2827 & 0.2054 & 0.8457 \\
+ Geometry Regularization  & ~ & \textbf{0.03303} & 0.2828 & 0.2053 & \textbf{0.8481} \\
+ Warping Pseudo~(Ours) & ~ & 0.03399 & \textbf{0.2796} & \textbf{0.2025} & 0.8469 \\
\hline

$\mathcal{W}^+$ Inversion & \multirow{5}*{L60} & 0.08951 & 0.4200 & 0.3485 & 0.7421 \\
+ Symmetry Prior & ~ & 0.04824 & 0.3251 & 0.2633 & 0.8202 \\
+ Joint Optimization  & ~ & 0.04087 & 0.3144 & 0.2424 & 0.8270 \\
+ Geometry Regularization  & ~ & \textbf{0.04032} & 0.3134 & 0.2416 & \textbf{0.8281} \\
+ Warping Pseudo~(Ours) & ~ & 0.04069 & \textbf{0.3113} & \textbf{0.2379} & 0.8272 \\

\hline

$\mathcal{W}^+$ Inversion & \multirow{5}*{R30} & 0.05888 & 0.3478 & 0.2938 & 0.7987 \\
+ Symmetry Prior & ~ & 0.03825 & 0.3013 & 0.2421 & 0.8244 \\
+ Joint Optimization  & ~ & 0.03133 & 0.2820 & 0.2083 & 0.8455 \\
+ Geometry Regularization  & ~ & \textbf{0.03134} & 0.2817 & 0.2081 & 0.8471 \\
+ Warping Pseudo~(Ours) & ~ & 0.03203 & \textbf{0.2807} & \textbf{0.2057} & \textbf{0.8529} \\

\hline
$\mathcal{W}^+$ Inversion & \multirow{5}*{R60} & 0.09239 & 0.4229 & 0.3587 & 0.7461 \\
+ Symmetry Prior & ~ & 0.05352 & 0.3361 & 0.2744 & 0.8140 \\
+ Joint Optimization  & ~ & 0.04565 & 0.3166 & 0.2465 & 0.8329 \\
+ Geometry Regularization  & ~ & \textbf{0.04488} & 0.3161 & 0.2448 & 0.8307 \\
+ Warping Pseudo~(Ours) & ~ & 0.04541 & \textbf{0.3160} & \textbf{0.2400} & \textbf{0.8335} \\

\hline
\end{tabular*}
\caption{Quantitative comparison on MEAD~\cite{kaisiyuan2020mead}.
View denotes the yaw angle of the input image. F is frontal, L is left side, and R is right side. 30 and 60 are the rotation degrees.
The metrics are calculated between the ground truth and the synthetic images in different views.
}
\label{tb:mead_ablation_module}
\vspace{-2pt}
\end{table*}

%% file: image_tex/img_ablation_appendix.tex
\begin{figure*}[t]
\begin{center}
\centerline{\includegraphics[width=0.95\linewidth]{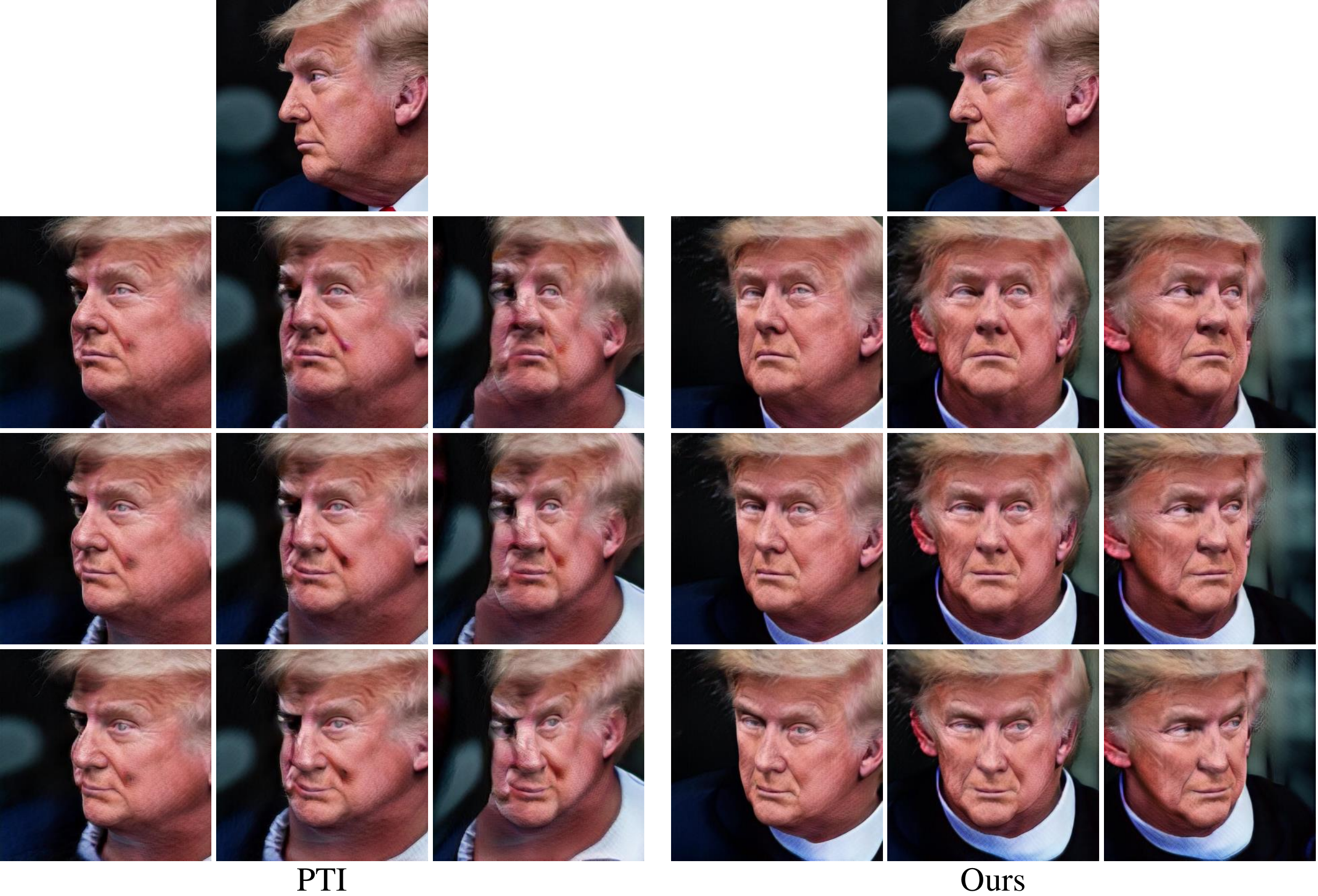}}
\centerline{\includegraphics[width=0.95\linewidth]{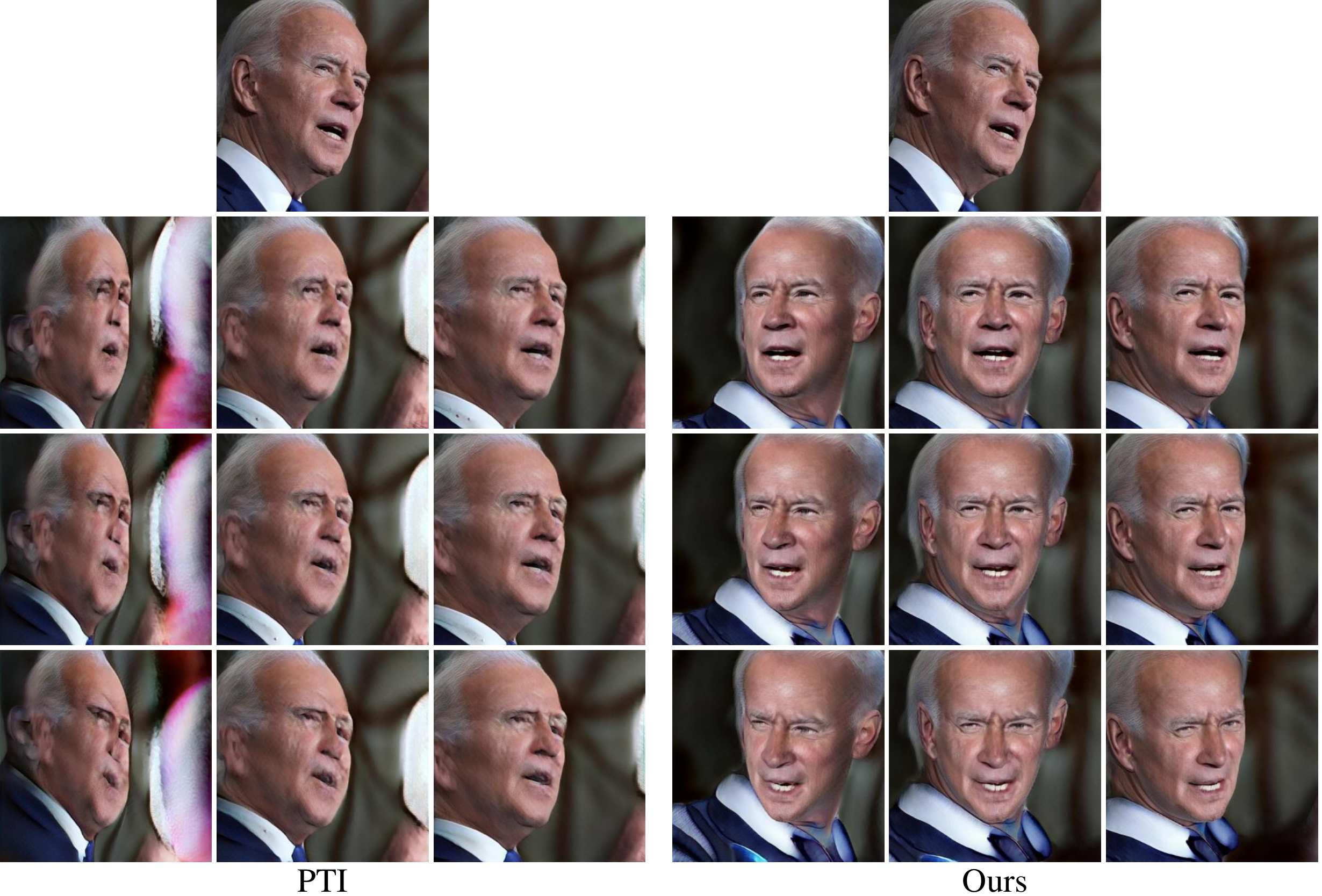}}
\caption{
Qualitative comparisons with PTI~\cite{roich2021PTI} on in-the-wild images.
}
\label{fig:appendix_1}
\vspace{-1cm}
\end{center}
\end{figure*}

\begin{figure*}[t]
\begin{center}
\centerline{\includegraphics[width=0.95\linewidth]{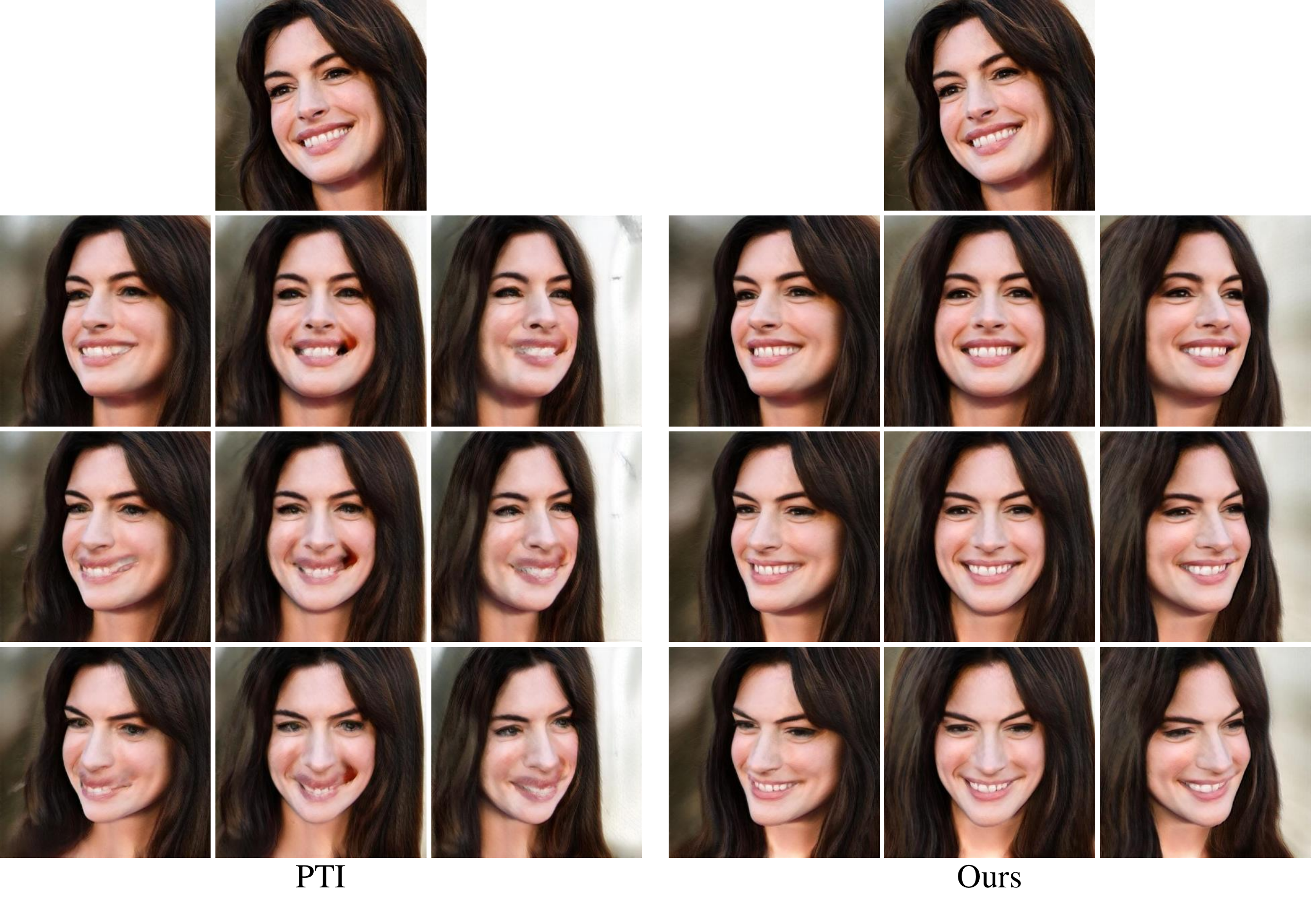}}
\centerline{\includegraphics[width=0.95\linewidth]{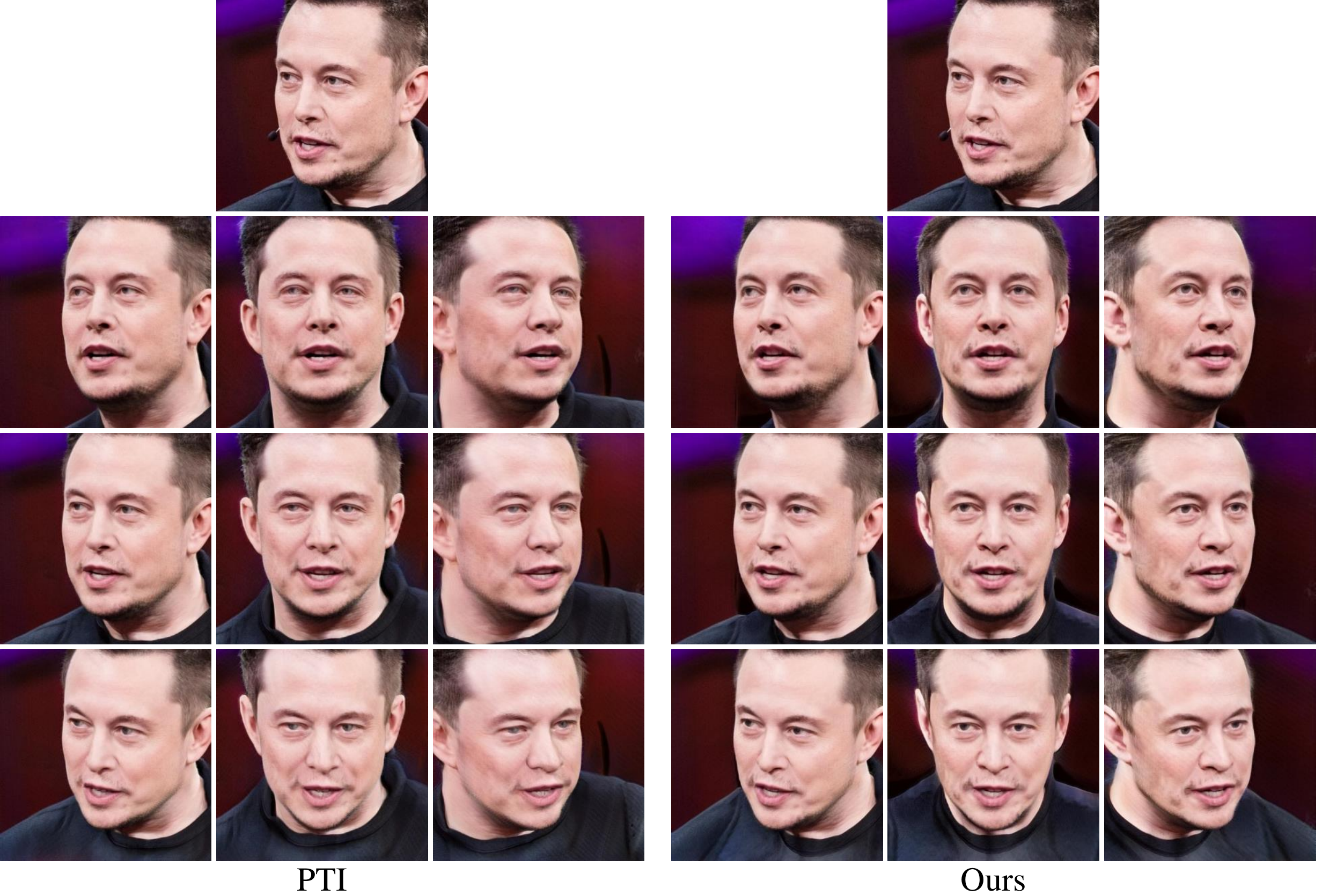}}
\caption{
Qualitative comparisons with PTI~\cite{roich2021PTI} on in-the-wild images.
}
\label{fig:appendix_2}
\vspace{-1cm}
\end{center}
\end{figure*}